\documentclass{article}

\usepackage[preprint]{neurips_2024}

\usepackage{natbib}
\setcitestyle{numbers,square}
\usepackage[utf8]{inputenc} 
\usepackage[T1]{fontenc}    
\usepackage{hyperref}       
\usepackage{url}            
\usepackage{booktabs}       
\usepackage{amsfonts}       
\usepackage{nicefrac}       
\usepackage{microtype}      
\usepackage{graphicx}
\usepackage{multirow}
\usepackage{wrapfig}
\usepackage{makecell}       
\usepackage{booktabs} 
\usepackage{soul} 
\usepackage{color, xcolor}

\title{MindSemantix: Deciphering Brain Visual Experiences with a Brain-Language Model}

\author{Ziqi Ren\\
    Xidian University\\
    \And
    Jie Li\\
    Xidian University\\
    \And 
    Xuetong Xue\\
    Xidian University\\
    \And 
    Xin Li\\
    Group 42 (G42)\\
    \AND 
    Fan Yang\\
    Group 42 (G42)\\
    \And
    Zhicheng Jiao\\
    Brown University\\
    Rhode Island Hospital\\
    \And
    Xinbo Gao\\
    Xidian University\\
    xbgao@mail.xidian.edu.cn\\}

\begin{document}
\bibliographystyle{plain}

\maketitle

\begin{abstract}
Deciphering the human visual experience through brain activities captured by functional Magnetic Resonance Imaging (fMRI) represents a compelling and cutting-edge challenge in the field of neuroscience research. Compared to merely predicting the viewed image itself, decoding brain activity into meaningful captions provides a higher-level interpretation and summarization of visual information, which naturally enhances the application flexibility in real-world situations. In this work, we introduce MindSemantix, a novel multi-modal framework that enables Large Language Models (LLMs) to comprehend visually-evoked semantic content in brain activity. Our MindSemantix explores a more ideal brain captioning paradigm by weaving LLMs into brain activity analysis, crafting a seamless, end-to-end Brain-Language Model. To effectively capture semantic information from brain responses, we propose Brain-Text Transformer, utilizing a Brain Q-Former as its core architecture. It integrates a pre-trained brain encoder with a frozen LLM to achieve multi-modal alignment of brain-vision-language and establish a robust brain-language correspondence. To enhance the generalizability of neural representations, we pre-train our brain encoder on a large-scale, cross-subject fMRI dataset using self-supervised learning techniques. MindSemantix provides more feasibility to downstream brain decoding tasks such as stimulus reconstruction. Conditioned by MindSemantix captioning, our framework facilitates this process by integrating with advanced generative models like Stable Diffusion and excels in understanding brain visual perception. MindSemantix generates high-quality captions that are deeply rooted in the visual and semantic information derived from brain activity. This approach has demonstrated substantial quantitative improvements over prior art. Our code will be publicly released at \url{https://github.com/ziqiren/MindSemantix}.
\end{abstract}

\section{Introduction}

Human brains persistently perceive, process and interpret various external stimuli, not only receiving the information of stimulus itself, but naturally constructing semantic understanding through the more complex brain regions such as temporal and frontal lobes\cite{popham2021visual}. Benefit from such brain mechanism, when perceiving visual stimuli, the neural signals measured with functional Magnetic Resonance Imaging (fMRI) have also demonstrated a potential to be decoded into text modality besides the image modality in previous \cite{beliy2019voxels,fang2020reconstructing,gaziv2022self,mozafari2020reconstructing,ren2021reconstructing,shen2019end,shen2019deep,lin2022mind,takagi2023high,ren2024brain,ren2023reconstructing}, which derives an emerging task, ``brain captioning'', and draws great attention\cite{ferrante2023multimodal,takagi2023improving,mai2023unibrain,chatterjee2023dreamcatcher}. Brain captioning refers to transform the semantic understanding captured in brain signals into natural languages via decoding model to describe the observed images. As the abstract summarization of visual content, captioning could facilitate visual decoding by approximatively simulating the reverse process of perceiving stimuli, low-level visual processing and high-level semantic processing in human brain\cite{binder2009semantic}. It comes more conforming to how the brain processes and represents complex visual information than the direct mapping from fMRI to vision. The integration of language and vision may revolutionize our understanding of the neural code underlying visual perception, and will be more flexible in possible applications in brain-computer interfaces\cite{gao2021interface,vansteensel2020brain,verbaarschot2021visual} and clinical diagnostics\cite{lu2023visualizing,struijs2021psychological} than using single image modality. 

Although the preliminary developing of decoding models for brain captioning has occurred, these works mapped fMRI patterns to the embeddings of text generation models via relatively simple mappings, usually ridge regression\cite{takagi2023improving,ferrante2023multimodal,mai2023unibrain}. Recent neuroscience research has shown the evidence of possible mechanism correlation between Large Language Models (LLMs) and human brains\cite{hardy2023large,mahowald2024dissociating,sejnowski2023large,jamali2023unveiling}, and attempts in other data modalities, like video and audio, have proved the feasibility of enabling LLMs to understand multi-modal content\cite{zhang2023video,chu2023qwen,wu2023multimodal}. Here we propose MindSemantix, a novel framework that empowers LLM with the capability of understanding visually-relevant semantic content in the brain activity and establishes an end-to-end Brain-Language Model (BLM) to achieve state-of-the-art brain captioning via fMRI. MindSemantix innovatively allows LLM to participate into the brain activity comprehension, unlike merely serving for text generation stage as in previous works\cite{takagi2023improving,ferrante2023multimodal}.

\begin{figure*}
  \centering
       \includegraphics[scale=.3]{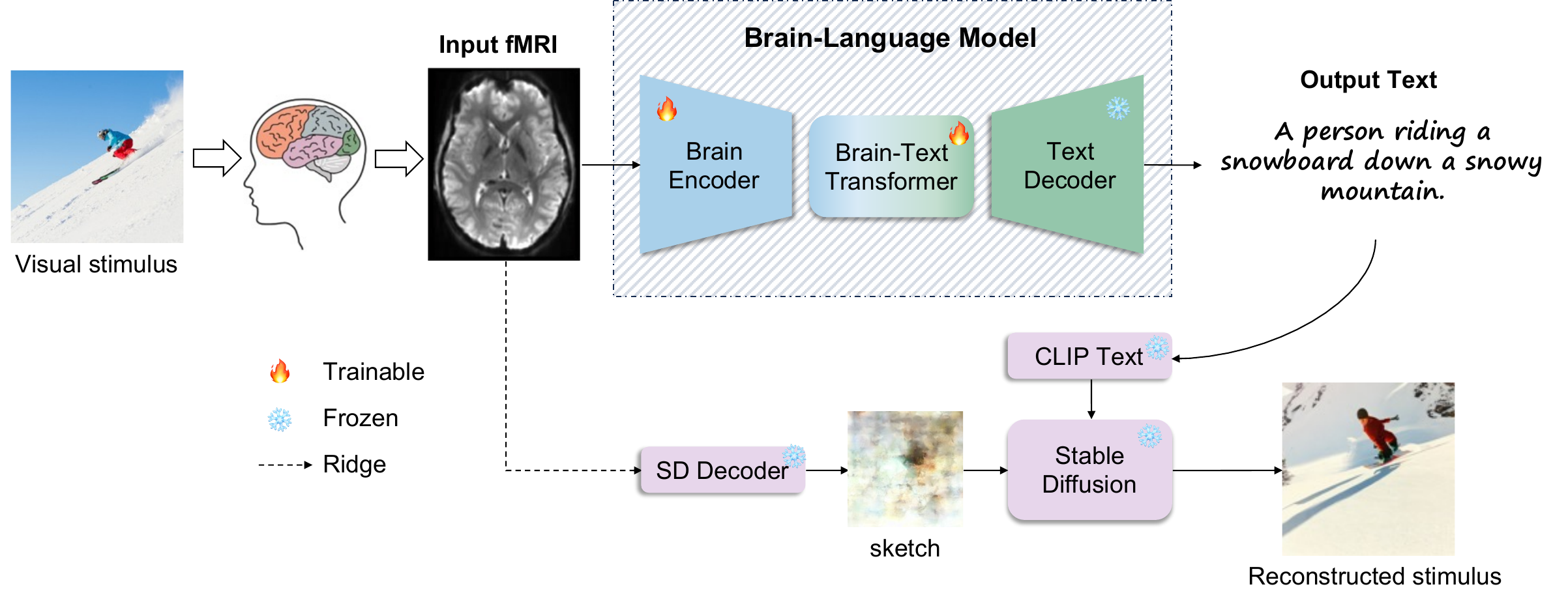}
  \caption{MindSemantix overall schematic.}
  \label{FIG:2}
\end{figure*}

BLM in our MindSemantix consists of three modules (see Figure\ref{FIG:2}), Brain Encoder, Brain-Text Transformer, and Text Decoder within the frozen Open Pretrained Transformer (OPT)\cite{zhang2022opt} as backbone. We pre-train Brain Encoder using large-scale cross-subject fMRI data via self-supervised learning. MindSemantix then bootstraps cross-modal training from the pre-trained Brain Encoder and the frozen OPT. Different from other maturer data modalities (\emph{e.g.}, image, text, video, \emph{etc.}), fMRI data suffer from the limitation of sample scarcity and low signal-noise-ratio (SNR), the challenge of multi-modal alignment greatly increasing. To solve this problem, we devise Brain-Text Transformer adopting a Brain Q-Former and design a training strategy inspired by the idea of BLIP-2\cite{li2023blip} to guarantee the training efficiency. MindSemantix captions can easily provide a semantic guidance to support realistic stimulus recovering through Stable Diffusion\cite{rombach2022high}.

Our contributions are as follows: (1) to the best of our knowledge, this is the first work that constructs an end-to-end decoding model specialized for brain captioning via fMRI. (2) Engaging LLMs into the brain activity comprehension, not just limited to the text generation in previous, fills in the blank of empowering LLMs with the capability of understanding brain visual perception. (3) State-of-the-art brain captioning performance can be obtained, quantitatively demonstrated on both low-level and high-level text metrics. (4) A convenient platform for potential downstream tasks like stimulus reconstruction is presented by flexibly assembling with advanced generation models.

\section{MindSemantix}

\begin{figure}
  \centering
       \includegraphics[scale=.46]{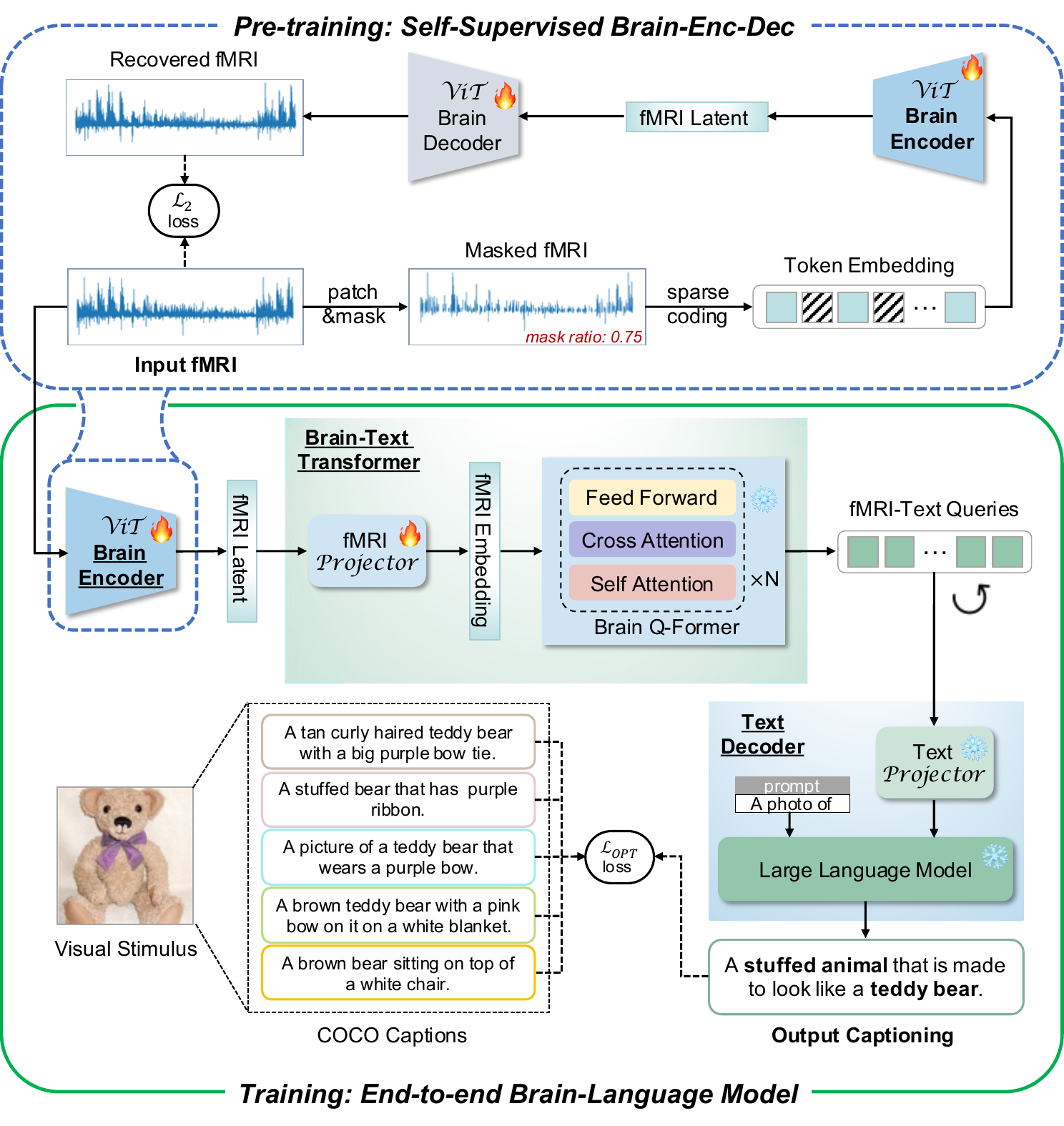}
  \caption{Learning procedure of MindSemantix. Top: pre-training phase of self-supervised Brain-Encoder-Decoder. Bottom: training phase of end-to-end Brain-Language Model.}
  \label{FIG:3}
\end{figure}

MindSemantix learns a seamless Brain-Language Model using an end-to-end trainable approach to decode fMRI patterns of specific subject into text modality (see Figure\ref{FIG:3}). By weaving a frozen Large Language Model (LLM) into brain activity comprehension, we map fMRI patterns into a well pre-trained visual-language embedding space. It delicately achieves a multi-modal alignment of brain-vision-language and is the core innovation of our method. To capture generalizable neural representations, we model an extra encoder-decoder architecture to pre-train the brain encoder by self-supervised learning technique using large-scale fMRI data collected from diverse subjects\cite{allen2022massive}. With the decoded captions as condition, MindSemantix can serve for visual stimulus reconstruction by integrated with Stable Diffusion. Further details will be provided in the subsequent subsections.

\subsection{Pre-training: Self-supervised Brain-Encoder-Decoder (BED)}

fMRI measures blood-oxygen-level-dependent (BOLD) changes as 3D voxels to record neuronal activities in human brain. The measured neuronal activities can be analyzed hierarchically with voxel-comprised functional networks which have implicit correlations with each other in response to external stimuli\cite{cordes2002hierarchical,wang2013analysis,van2010exploring,zhou2010divergent}. In addition, the hemodynamic response and spatial smoothing functions in fMRI BOLD signal jointly cause spatial blurring, creating spatial redundancy in fMRI data and similar amplitudes between neighboring voxels\cite{engel1997retinotopic,shmuel2007spatio,uugurbil2013pushing}. Considering these analyses altogether, we introduce a masked auto-encoder (MAE)\cite{chen2023seeing} approach termed as ``Self-supervised Brain-Encoder-Decoder" to capture valuable information of neuronal activities from fMRI voxels. 

The vectorized input voxels are divided into patches which are then randomly masked with an extremely high mask ratio (75\%) following \cite{chen2023seeing} to save computations without losing the learning power of masked modeling. The masked fMRI patterns are subsequently tokenized into embeddings using a 1D convolutional layer with a stride equal to the patch size. Based on the sparse coding mechanism in human brain\cite{lazebnik2006beyond}, we employ a large embedding-to-patch-size ratio to increase information capacity with a large fMRI latent space. BED adopts ViT-Large\cite{dosovitskiy2020image} structure, in which the encoder serves for learning effective fMRI representations, while the decoder used to predict the masked patches will be discarded as long as the pre-training converges. 

To efficiently endows BE with the capability of representation generalizability across subjects, we train BED using large-scale, dimensional-aligned fMRI data of diverse subjects with $\mathcal{L}_{2}$ loss:
\begin{equation}
\mathcal{L}_{BED} = \sum_{i=1}^{N_{all}}\mathcal{L}_{2}(\mathbf{x}'_{i}, \mathbf{x}_{i})=\sum_{i=1}^{N}\left\| \mathbf{x}_{i}-\mathbf{x}'_{i}\right\|_{2}
\end{equation}
where $\mathbf{x}_{i}$ and $\mathbf{x}'_{i}$ represent $i$-th fMRI sample and its recovered fMRI pattern respectively. $N_{all}$ is the total data size of fMRI samples involving all subjects.

\subsection{Training: End-to-end Brain-Language Model (BLM)}

We propose a general and efficient Brain-Language Model for brain captioning from human fMRI activity. As illustrated in Figure\ref{FIG:3}, BLM consists of Brain Encoder (BE), Brain-Text Transformer (BT-Former) and Text Decoder (TD), serving for brain representation learning, multi-modal alignment of brain and language, and text generation, respectively. 

Compared with the data modality of image or text, fMRI data always suffer from the limitation of sample scarcity and complex measuring noise, causing an inherent obstacle for bridging the gap between fMRI and perception stimuli. To overcome this problem, we design BT-Former containing a fMRI Projector (${P}_{fMRI}$) and a Brain Q-Former. The structure of Brain Q-Former is a transformer similar to the Q-Former module in BLIP-2\cite{li2023blip}, a multi-modal Transformer model for visual-language pre-training (VLP). We initialize Brain Q-Former with the BLIP-2 model pre-trained on large-scale data for image captioning task, which offers a free prior of visual-language alignment embedding space, and keep it frozen during training process. To bind brain patterns to the visual-language prior space, we use ${P}_{fMRI}$ composed of a fully-connected (FC) layer and a 1D convolutional layer to project the output of BE into the same dimension as the input embedding of Brain Q-Former. Thus, BT-Former transforms fMRI features into a series of trainable queries corresponding to relevant visual and semantic information.

TD is established on the basis of a frozen LLM, here we experiment on OPT-2.7B\cite{zhang2022opt}. To harvest the LLM's generative language capability, a Text Projector (${P}_{text}$) is designed to connect BT-Former and the LLM, which linearly projects the output fMRI-text query embeddings into the same dimension as the text embedding of LLM. The projected query embeddings are then prepended to the input text embeddings. Since the initialized BT-Former could extract language-informative brain representation, it effectively functions as an information bottleneck that feeds the most useful information to the LLM while removing irrelevant brain information. The frozen LLM then is capable of generating text conditioned on the brain representation learned from visual-evoked fMRI. In addition, we structure ${P}_{text}$ basically like the Fully-Connected module in BLIP-2 to share its pre-trained weights, and keep frozen during training. 

Therefore, we learn the whole BLM with the language modeling loss of OPT ($\mathcal{L}_{OPT}$) between the generated brain captions and the ground-truth COCO captions, while only BE and ${P}_{fMRI}$ are actually trained:
\begin{equation}
\mathcal{L}_{BLM} = \sum_{i=1}^{N}\mathcal{L}_{OPT}(\mathbf{c}'_{i}, \mathbf{c}_{i})=\sum_{i=1}^{N}\left [ \sum_{j=1}^{M}\mathcal{L}_{OPT}(\mathrm{BLM}(\mathbf{x}_{i}),\mathbf{c}_{ij})\right ] 
\end{equation}
where $\mathbf{c}_{i}$ and $\mathbf{c}'_{i}$ represent true stimulus captions and the predicted caption of $i$-th fMRI sample, respectively. $M$ denotes the number of COCO captions corresponding to each stimulus image, and $M=5$. We use all the presented COCO captions to train our BLM to enhance the semantic richness and structural flexibility of text description. $N$ is the data size of training fMRI samples for specific subject.

\subsection{MindSemantix for Visual Reconstruction}

For complex visual stimuli, the contained abundant information cause it quite difficult to accurately characterize the stimulus images from brain activity, so we make MindSemantix captions as a prior to guide visual reconstruction. As depicted in Figure\ref{FIG:2}, realistic, high-resolution reconstructions can be synthesized through Stable Diffusion\cite{rombach2022high} (SD) by embedding our brain captions.

SD first learns visual representation (denoted as $\mathbf{Z}_{vis}$) from the stimulus image $\mathbf{Y}$ using an auto-encoder trained on a very large-scale image dataset. We train a ridge regressor to linearly map the fMRI pattern $\mathbf{X}$ to $\mathbf{Z}_{vis}$ and decode to sketch (denoted as $\mathbf{Y}'_{vis}$) representing the layout of low-level visual information in image such as structure and color. $\mathbf{Y}'_{vis}$ is generated via the decoder module in SD and can be regarded as a basic guess for the final reconstruction. Then $\mathbf{Y}'_{vis}$ is compressed into the latent space via the encoder module in SD and gradually incorporated with Gaussian noise through the forward diffusion process to destroy the structure of data. The noisy variant of the compressed latent input $\mathbf{z}$ at each time point is defined as $\mathbf{z}_{t}=\sqrt{\alpha_{t}}\mathbf{z}+\sqrt{1-\alpha_{t}}\epsilon _{t}$, where $t\in \left \{ 1,\dots,T \right \}$, $\alpha$ is a hyperparameter, and $\epsilon$ is the Gaussian. In the reverse diffusion process, MindSemantix captioning ($\mathbf{C}'$) is compressed into semantic embedding through the pre-trained text encoder in CLIP\cite{radford2021learning} and inserted into the denoising U-Net\cite{ronneberger2015u} via cross-attention. The final latent representation $\mathbf{z}_{0}$ derived after all diffusion steps is given as input to SD Decoder to produce high-resolution reconstruction.

\section{Experimental Results}

\subsection{Dataset and Setting}

\textbf{Natural Scenes Dataset (NSD)} 
We conducted experiments on the publicly accessible NSD\cite{allen2022massive} dataset, containing high-resolution 7-Tesla fMRI scans captured from 8 subjects viewing images from the COCO\cite{lin2014microsoft} dataset. We focus on 4 subjects (Sub-1, Sub-2, Sub-3, and Sub-7) who finished all viewing trials. Each subject’s training set includes 8859 image stimuli and 24980 fMRI trials, while the test set comprises 982 image stimuli and 2770 fMRI trials, with stimuli differing across subjects in the training set but shared in the test set. The average of the three trials associated with each image was used for the test set, but the training set employed the separate trials without averaging. Utilizing the NSDGeneral region-of-interest (ROI) mask at 1.8 mm resolution, we derived ROI data encompassing the following voxel counts for the 4 subjects: [15724, 14278, 13039, 12682]. Corresponding captions can be extracted from the COCO dataset.

\textbf{Implementation Details}
Our MindSemantix is implemented on NVIDIA RTX 3090 Ti GPU. In the pre-training phase, BDE is build as an asymmetric auto-encoder architecture following \cite{chen2023seeing}, in which the decoder is considerably smaller with 8 layers than the encoder with 24 layers. Voxel patch size is set as 16 and the dimension of token embedding is 1024. We make an alignment dimension of cross-subject fMRI data as 15728 according to the patch size. We pre-train BDE for 500 epochs with 40-epoch warming up, an initial learning rate of 2.5e-4, and a batch size of 16. In the training phase, BLM is trained for 10 epochs with 1000-step warming up, the initial learning rate of 1e-5, and the batch size of 2. We employ AdamW\cite{loshchilov2017decoupled} for optimization with a weight decay of 0.05 during both pre-training and training phase. For visual reconstruction, we use 50 diffusion steps with a diffusion strength of 0.8 in the diffusion process. More details will be found in our code.

\subsection{Evaluation Metric}

We introduce various text evaluation metrics in both low-level and high-level aspects as follows to make comprehensive quantitative comparisons with current models.

\textbf{Low-level Metric}
Low-level text features provide basic information about the structure and composition of the text, measured by: i) \textbf{Meteor}: Meteor\cite{banerjee2005meteor} provides a robust evaluation by considering word overlap, word order, synonymy, and other linguistic aspects. ii) \textbf{Rouge}: Rouge-L\cite{lin2004rouge}, a variant of the Rouge (Recall-Oriented Understudy for Gisting Evaluation) metric that focus on capturing the similarity between the generated and reference summaries. iii) \textbf{CIDEr}: An image captioning metric, based on the concept of consensus and taking into account both the language and content\cite{vedantam2015cider}.

\textbf{High-level Metric}
High-level text features are more complex and meaningful representations of text data that capture the context, relationships, and semantics of words and sentences, measured by: i) \textbf{SPICE}: A semantic evaluation metric measuring how effectively generated captions recover objects, attributes and the relations between them\cite{anderson2016spice}. ii) \textbf{CLIP}: CLIP similarity score, assessing how well the generated text aligns semantically with a reference text through the output layer of the CLIP-Text\cite{radford2021learning} model. iii) \textbf{Sentence}: SentenceTransformer similarity score\cite{reimers2019sentence}.

\begin{figure}
  \centering
       \includegraphics[scale=.328]{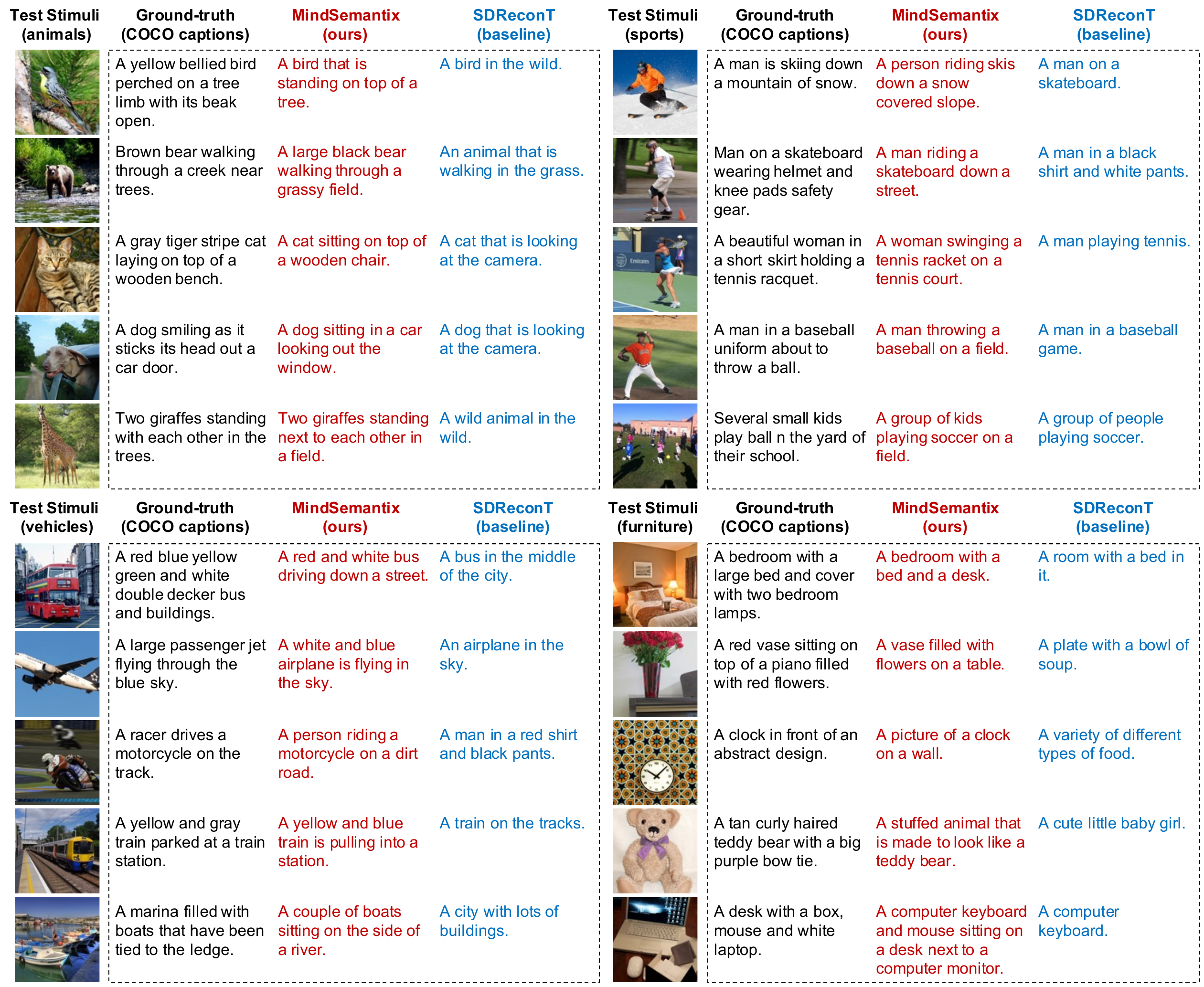}
  \caption{Sample brain captioning results across visual stimulus categories of MindSemantix and SOTA method. The same test set was used.}
  \label{FIG:6}
\end{figure}

\subsection{Captioning Results}

\begin{wrapfigure}{r}{0.5\textwidth}
\centering
\includegraphics[width=7cm]{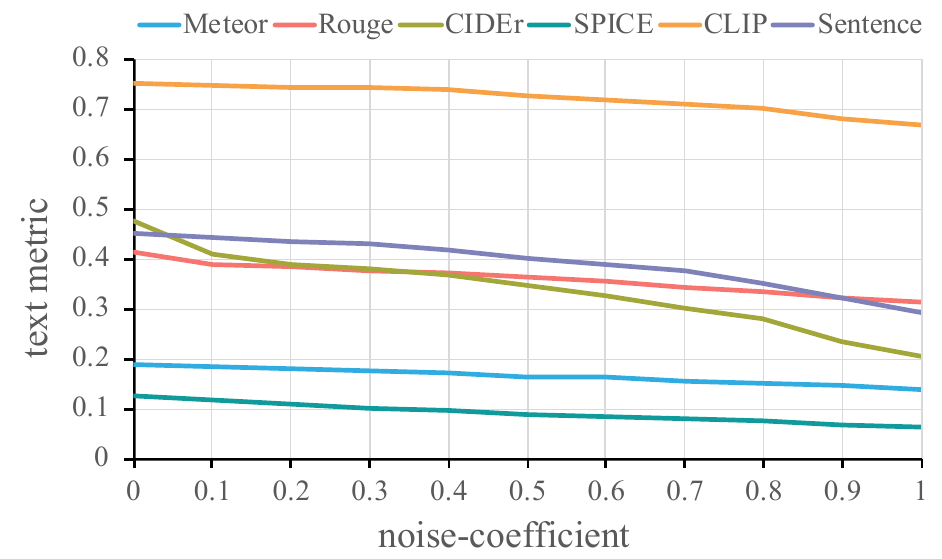}\
\caption{MindSemantix captioning performance on Subject1 fMRI added with Gaussian noise ($mean=0$, $std$ equals the mean of fMRI signal values) by coefficients from 0.1 to 1.}
\label{FIG:8}
\end{wrapfigure}

In this section, we present a comparative analysis of MindSemantix captioning results with preceding state-of-the-art methods including SDReconT\cite{takagi2023improving}, UniBrain\cite{mai2023unibrain} and BrainCap\cite{ferrante2023multimodal}, in which SDReconT serves as baseline. Table\ref{TAB:1} provides the quantitative evaluation results on Subject1 (Sub-1) and has demonstrated that MindSemantix outperforms all state-of-the-arts by a significant margin on all metrics. SDReconT\cite{takagi2023improving} first constructs a brain captioning pipeline composed of a linear feature regression model and a captioning model, while its performance pretty suffers from its limited vocabulary and a lot of repetitive words which are redundant or meaningless often appear in its captioning. UniBrain\cite{mai2023unibrain} builds two linear regression models to derive both low-level and high-level text latent representations, which obviously improves the fluency of generated sentences. BrainCap\cite{ferrante2023multimodal} follows a similar pipeline to baseline but replaces the captioning model, which acts better in captioning accuracy. However, these above methods are actually lack of direct links between brain signals and their corresponding captions. In contrast, MindSemantix decodes more accurate semantics and captures more sufficient information from brain signals by leveraging all the presented COCO captions to provide a direct guidance, which is a free and effective strategy. Moreover, the sufficient participation of LLM in MindSemantix makes it not only excel in perceiving and understanding neural representations, but also generate sentences that are fluent, complete, and rich in vocabulary. As an approximate upper bound, we also report `BLIP-2/ImgCap' which utilizes BLIP-2\cite{li2023blip} for captioning from the ground truth images (visual stimuli). Moreover, we verified that MindSemantix performs a strong robustness when decoding the test fMRI signals added with different degrees of Gaussian noise. As shown in Figure\ref{FIG:8}, the reduction of captioning performance on most metrics is quite slight with noise increasing.

\begin{table}
  \caption{Compare quantitative results of MindSemantix against SOTA methods on Subject1. The \textbf{best} scores are BOLD.}
  \label{TAB:1}
  \centering
  \setlength{\tabcolsep}{8pt} 
  \renewcommand{\arraystretch}{1.2} 
  \footnotesize
  \begin{tabular}{ccccccc}
    \toprule[1pt]
    \multirow{2}{*}{Method} &\multicolumn{3}{c}{Low-Level}&\multicolumn{3}{c}{High-Level}\\ 
    \cmidrule(lr){2-4}\cmidrule(lr){5-7}
			&Meteor $\uparrow$ & Rouge $\uparrow$ & CIDEr $\uparrow$ & SPICE $\uparrow$ & CLIP $\uparrow$ & Sentence $\uparrow$ \\
\hline
BLIP-2/ImgCap\cite{li2023blip} & 0.330 & 0.591 & 1.264 & 0.265 & 0.911 & 0.827 \\
\hline
SDReconT\cite{takagi2023improving} & 0.100 & 0.251 & 0.138 & 0.050 & 0.624 & 0.280 \\
UniBrain\cite{mai2023unibrain} & 0.169 & 0.222 & -- & -- & -- & -- \\
BrainCap\cite{ferrante2023multimodal} & 0.167 & 0.407 & 0.413 & 0.091 & 0.705 & 0.447 \\
\hline	
MindSemantix & \bfseries{0.190} & \bfseries{0.415}  & \bfseries{0.476} & \bfseries{0.125} & \bfseries{0.755} & \bfseries{0.454}\\
\bottomrule[1pt]
  \end{tabular}
\end{table}

Since UniBrain\cite{mai2023unibrain} and BrainCap\cite{ferrante2023multimodal} have not been open-sourced, here we make qualitative analysis on MindSemantix and SDReconT\cite{takagi2023improving}. As shown in Figure\ref{FIG:6}, we presented the captioning results of MindSemantix and SDReconT for different categories of visual stimuli, involving animal, sport, vehicle, and furniture. From a comprehensive perspective, our model performs a remarkable superiority over the baseline in both semantic accuracy and information completeness. Especially, MindSemantix captioning results preserve a more fine-grained consistency with the ground-truth. For example, MindSemantix precisely describes the objective in the second animal sample as `a large black bear' while SDReconT just provides a rough description of `an animal', and `a yellow and blue train' of MindSemantix for the forth vehicle sample provides detailed information of color while SDReconT misses. Moreover, MindSemantix vividly depicts the details of human behaviour in the viewed sport stimuli besides the objective and place, such as 'a woman swinging a tennis racket on a tennis court' for the third sample. Those rich caption details will provide a solid basis for the accurate reconstruction in downstream.

\subsection{MindSemantix for Reconstruction}

\begin{figure}[h]
  \centering
       \includegraphics[scale=.285]{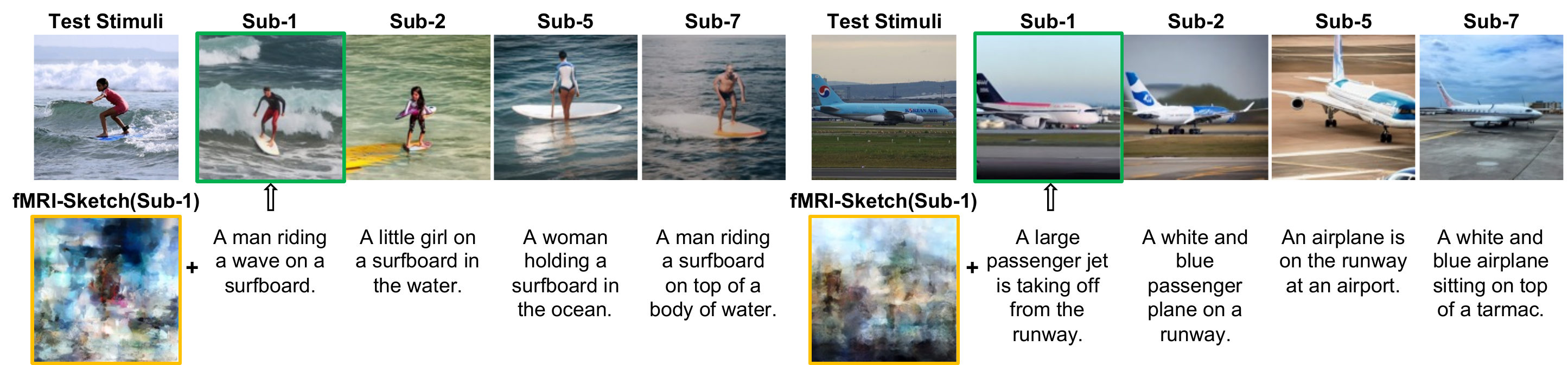}
  \caption{Sample visual reconstruction and captioning results of MindSemantix on each subject.}
  \label{FIG:5}
\end{figure}

\begin{figure}
  \centering
       \includegraphics[scale=.5]{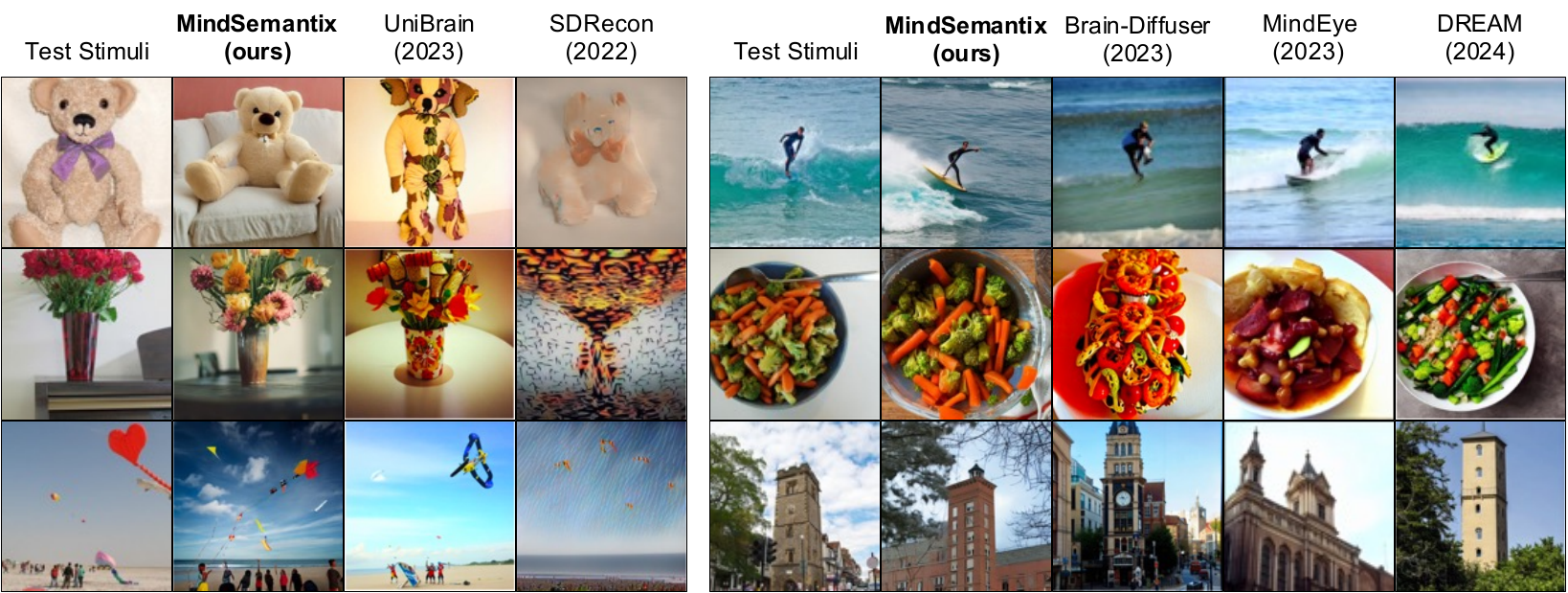}
  \caption{Sample visual reconstruction results from MindSemantix and SOTA methods on NSD\cite{allen2022massive}. The same test set was used across methods.}
  \label{FIG:7}
\end{figure}

With predicted captions, we reconstruct visual stimuli based on the brain-decoded sketches utilizing Stable Diffusion\cite{rombach2022high}. Figure\ref{FIG:5} presents some samples of generated images from fMRI and their corresponding MindSemantix captions. The reconstructed images of different subjects evidently preserve most of the layout and semantics presented in visual stimuli, which demonstrates a superior generalization performance of our model. Semantics in the reconstructions are highly-consistent with their corresponding brain captions, while sometimes slightly vary across subjects, such as `a man' of Sub-1 and `a little girl' of Sub-2 for the first sample, as reflected in their reconstructed images. 

We contrast the qualitative results of MindSemantix with current state-of-the-art methods\cite{takagi2023high,ozcelik2023natural,mai2023unibrain,scotti2024reconstructing,xia2024dream} in Figure\ref{FIG:7}. We chose distinct images for each model to facilitate comparison, considering the variations in test images presented across their respective papers. SDRecon\cite{takagi2023high} serves as baseline in this task, which can produce identifiable silhouettes but falls short in terms of many qualitative aspects (\emph{i.e.}, low-level details, high-level semantics, and naturality). The subsequent methods\cite{ozcelik2023natural,mai2023unibrain,scotti2024reconstructing,xia2024dream} have effectively improved these aspects, and our model exhibits an obvious superiority in semantics and naturality, such as the generated teddy bear, flowers, and food of MindSemantix performing more realistic than other models. Since without more specific designs tailored for this task, our resulting images are not particularly rigorous in spatial details. Table~\ref{TAB:2} shows that our method quantitatively achieves comparably or better than state-of-the-art methods especially on high-level metrics.

\begin{table}
  \caption{Quantitative reconstruction comparison of SOTA methods average across the 4 subjects, except MindReader which only analyzed Subject1. The results were measured by image metrics including \textbf{PixCorr}: pixel-level correlation; \textbf{SSIM}: structural similarity index\cite{wang2004image}; \textbf{AlexNet}: AlexNet-2 and AlexNet-5, two-way identifications of the second and fifth layers of AlexNet\cite{krizhevsky2012imagenet}, respectively; \textbf{Inception}: two-way identification of the last pooling layer of InceptionV3\cite{szegedy2016rethinking}; \textbf{CLIP}: two-way identification of the output layer of the CLIP-Image\cite{radford2021learning} model; \textbf{EffNet}: a distance metric gathered from EfficientNet-B1\cite{tan2019efficientnet}; \textbf{SwAV}: a distance metric gathered from SwAV-ResNet50\cite{caron2020unsupervised}. The \sethlcolor{red!35}\hl{best}, \sethlcolor{orange!40}\hl{second} and \sethlcolor{yellow!50}\hl{third} scores are highlighted.}
  \label{TAB:2}
  \centering
  \setlength{\tabcolsep}{2.2pt} 
  \renewcommand{\arraystretch}{1.2} 
  \footnotesize
\begin{tabular}{ccccccccc}
\toprule[1pt]
			\multirow{2}{*}{Method} &\multicolumn{4}{c}{Low-Level}&\multicolumn{4}{c}{High-Level}\\
			\cmidrule(lr){2-5}\cmidrule(lr){6-9}
			&PixCorr $\uparrow$ & SSIM $\uparrow$ & AlexNet-2 $\uparrow$ & AlexNet-5 $\uparrow$ & Inception $\uparrow$ & CLIP $\uparrow$  & EffNet-B $\downarrow$ & SwAV $\downarrow$ \\
\hline
MindReader\cite{rombach2022high} & -- & -- & -- & -- & 78.2\% & --  & -- & --\\
SDRecon\cite{takagi2023high}  & -- & -- & 81.4\% & 81.5\%  & 76.0\% & 77.0\%  & -- & -- \\
BrainDiffuser\cite{ozcelik2023natural} & 0.254 & \sethlcolor{red!35}\hl{0.354} & \sethlcolor{orange!40}\hl{94.2\%} & 96.2\% & 87.2\% & 91.5\% & 0.775 & 0.423 \\
UniBrain\cite{mai2023unibrain} & 0.249 & \sethlcolor{yellow!50}\hl{0.330} & 92.9\% & 95.6\% & 87.8\% & 92.3\% & 0.766 & \sethlcolor{yellow!50}\hl{0.407} \\
MindEye\cite{scotti2024reconstructing} & \sethlcolor{red!35}\hl{0.309} & 0.323 & \sethlcolor{red!35}\hl{94.7\%} & \sethlcolor{red!35}\hl{97.8\%} & \sethlcolor{red!35}\hl{93.8\%}   & \sethlcolor{orange!40}\hl{94.1\%} & \sethlcolor{red!35}\hl{0.645} & \sethlcolor{red!35}\hl{0.367} \\
DREAM\cite{xia2024dream} & \sethlcolor{yellow!50}\hl{0.274} & 0.328 & \sethlcolor{yellow!50}\hl{93.9\%} & \sethlcolor{yellow!50}\hl{96.7\%}  & \sethlcolor{yellow!50}\hl{93.4\%} & \sethlcolor{orange!40}\hl{94.1\%} & \sethlcolor{red!35}\hl{0.645} & 0.418 \\
\hline	
MindSemantix & \sethlcolor{orange!40}\hl{0.299} & \sethlcolor{orange!40}\hl{0.333} & 93.1\% & \sethlcolor{orange!40}\hl{96.8\%} & \sethlcolor{red!35}\hl{93.8\%} & \sethlcolor{red!35}\hl{94.3\%}  & \sethlcolor{orange!40}\hl{0.686} & \sethlcolor{orange!40}\hl{0.392} \\
\bottomrule[1pt]
  \end{tabular}
\end{table}

\subsection{Ablation Study}

To verify the effectiveness of MindSemantix, we conducted ablation experiments on Subject1. We set ablation models based on the complete version of BLM’s settings to investigate the effect of each component, from the aspects of training strategy, model architecture, loss function, and scale of utilized COCO captions. \textbf{Training Strategy:} we removed the self-supervised pre-training of BED (SS-BED) to evaluate the impact of initialization on brain encoder. \textbf{Model Architecture:} we replaced our BE and BT-Former modules with a linear regression model similar to baseline\cite{takagi2023improving} to extract fMRI features. \textbf{Loss Function:} we replaced the language modeling loss ($\mathcal{L}_{OPT}$) between brain captions and COCO captions with the feature-level loss between the learned fMRI-text queries and the true image queries derived from BLIP-2 to train the model. \textbf{COCO-Caption Scale:} we trained the model using one or three randomly selected COCO captions of each training stimulus as ground-truth (denoted as $M=1$ and $M=3$). Quantitative measures for the ablation models have demonstrated the significance of each component and the complete BLM performs best (see Table\ref{TAB:3}).

We further examine the impact of MindSemantix captioning on visual reconstruction by evaluating the fMRI-derived sketches ($\mathbf{Y}'_{vis}$) and the images resulted through Stable Diffusion with $\mathbf{Y}'_{vis}$ fed in but no caption conditioned. Sufficient semantic information is injected into the final reconstructions via our predicted captions ($\mathbf{C}'$) and brings performance zooming especially on high-level metrics (see Table\ref{TAB:4}). Such phenomenon verified a strong influence of meaningful captions on reconstruction quality, which was scarcely concerned in most existing methods\cite{takagi2023high,ozcelik2023natural,scotti2024reconstructing,xia2024dream} and provides a promising inspiration for future optimization.

\begin{table}
  \caption{Ablation results of MindSemantix components on Subject1. In the complete BLM, $M=5$.}
  \label{TAB:3}
  \centering
  \setlength{\tabcolsep}{5pt} 
  \renewcommand{\arraystretch}{1.2} 
  \footnotesize
  \begin{tabular}{ccccccc}
    \toprule[1pt]
    \multirow{2}{*}{Model} &\multicolumn{3}{c}{Low-Level}&\multicolumn{3}{c}{High-Level}\\ 
    \cmidrule(lr){2-4}\cmidrule(lr){5-7}
			&Meteor $\uparrow$ & Rouge $\uparrow$ & CIDEr $\uparrow$ & SPICE $\uparrow$ & CLIP $\uparrow$ & Sentence $\uparrow$ \\
\hline	
BLM & \bfseries{0.190} & \bfseries{0.415}  & \bfseries{0.476} & \bfseries{0.125} & \bfseries{0.755} & \bfseries{0.454}\\
w/o SS-BED & 0.157 & 0.353 & 0.291 & 0.079 & 0.717 & 0.360 \\
w/o BE+BT-Former & 0.126 & 0.331 & 0.237 & 0.074 & 0.625 & 0.319 \\
w/o $\mathcal{L}_{OPT}$ & 0.156 & 0.355 & 0.322 & 0.095 & 0.754 & 0.431 \\
$M=1$ & 0.173 & 0.375 & 0.376 & 0.096 & 0.739 & 0.408 \\
$M=3$ & 0.184 & 0.384 & 0.408 & 0.112 & 0.749 & 0.437 \\
\bottomrule[1pt]
  \end{tabular}
\end{table}

\begin{table}
  \caption{Impact evaluation of MindSemantix on visual reconstruction using Subject1 data.}
  \label{TAB:4}
  \centering
  \setlength{\tabcolsep}{3pt} 
  \renewcommand{\arraystretch}{1.2} 
  \footnotesize
\begin{tabular}{ccccccccc}
\toprule[1pt]
			\multirow{2}{*}{Model} &\multicolumn{4}{c}{Low-Level}&\multicolumn{4}{c}{High-Level}\\
			\cmidrule(lr){2-5}\cmidrule(lr){6-9}
			&PixCorr $\uparrow$ & SSIM $\uparrow$ & AlexNet-2 $\uparrow$ & AlexNet-5 $\uparrow$ & Inception $\uparrow$ & CLIP $\uparrow$  & EffNet-B $\downarrow$ & SwAV $\downarrow$ \\
\hline
$\mathbf{Y}'_{vis}$ & \bfseries{0.354} & 0.286 & 74.7\% & 71.9\% & 54.0\% & 54.6\% & 0.993 & 0.720\\
$\mathbf{Y}'_{vis}$+SD & 0.318 & 0.295 & 74.7\% & 73.1\% & 57.1\% & 57.2\% & 0.974 & 0.628\\
$\mathbf{Y}'_{vis}$+SD+$\mathbf{C}'$ & 0.345 & \bfseries{0.331} & \bfseries{94.4\%} & \bfseries{98.5\%} & \bfseries{94.7\%} & \bfseries{95.8\%} & \bfseries{0.678} & \bfseries{0.384} \\
\bottomrule[1pt]
  \end{tabular}
\end{table}

\section{Discussion}

The experimental findings indicate that, MindSemantix can produce captions to effectively characterize brain visual experiences, substantially outperforming previous methodologies. Nonetheless, our analysis also uncovers some limitations within this work. Although we have used all the available COCO captions to train our BLM, there is a possible prospect to augment the ground-truth caption data by easily producing with advanced LLMs like GTP\cite{kalyan2023survey,achiam2023gpt} to elevate the training performance in future work. Additionally, besides the attempt of simple integration with Stable Diffusion has made in this work, the exploration of assembling MindSemantix with depth estimation, more powerful diffusion models, or the existing reconstruction-specialized methods may have a great potential to further improve stimulus reconstruction. The paradigm in MindSemantix could be expanded to other types of perception stimuli besides images and the exploration of multi-modal brain decoding system with language as a media will hold significant academic value and interest.

\section{Conclusion}

We present MindSemantix, a novel multi-modal framework that achieves state-of-the-art brain captioning of natural scene stimuli from fMRI activity. MindSemantix first introduces an end-to-end Brain-Language Model for decoding human visual experience, which revolutionizes the previous paradigm. In this Brain-Language Model, by integrating a frozen Large Language Model\cite{zhang2022opt} with a pre-trained brain encoder, a Brain-Text Transformer performs effective multi-modal alignment of brain-vision-language and learns a robust brain-language correspondence. Conditioned by MindSemantix captioning, we recovered realistic visual stimuli with high semantic fidelity through Stable Diffusion\cite{rombach2022high}. Extensive ablation studies verify the effectiveness of each proposed component.

\bibliography{neurips_2024}

\end{document}